\documentclass[10pt,twocolumn,letterpaper]{article}

\usepackage[pagenumbers]{cvpr}      % To produce the REVIEW version
\usepackage{multirow}
\usepackage[
singlelinecheck=false % <-- important
]{caption}
\usepackage{cuted}
\usepackage{stfloats}
\usepackage{lipsum}

\usepackage[dvipsnames]{xcolor}

\definecolor{cvprblue}{rgb}{0.21,0.49,0.74}
\usepackage[pagebackref,breaklinks,colorlinks,citecolor=cvprblue]{hyperref}

\def\paperID{38} 
\def\confName{CVPR}
\def\confYear{2024}

\title{The Curious Case of End Token: A Zero-Shot Disentangled Image Editing using CLIP}

\author{Hidir Yesiltepe \qquad
Yusuf Dalva \qquad
Pinar Yanardag \\
Virginia Tech\\
{\tt\small \{hidir, ydalva, pinary\}@vt.edu}
}

\begin{document}
\twocolumn[{
\maketitle    
    \begin{tabular}{c c c c c c c}
     \multicolumn{3}{c}{\qquad \qquad \small{} \qquad \qquad} & \multicolumn{2}{c}{\qquad \qquad \small{} \qquad \qquad } & \multicolumn{2}{c}{\qquad \qquad \small{} \qquad} \\
    \multicolumn{7}{c}{\includegraphics[width=1.\linewidth]{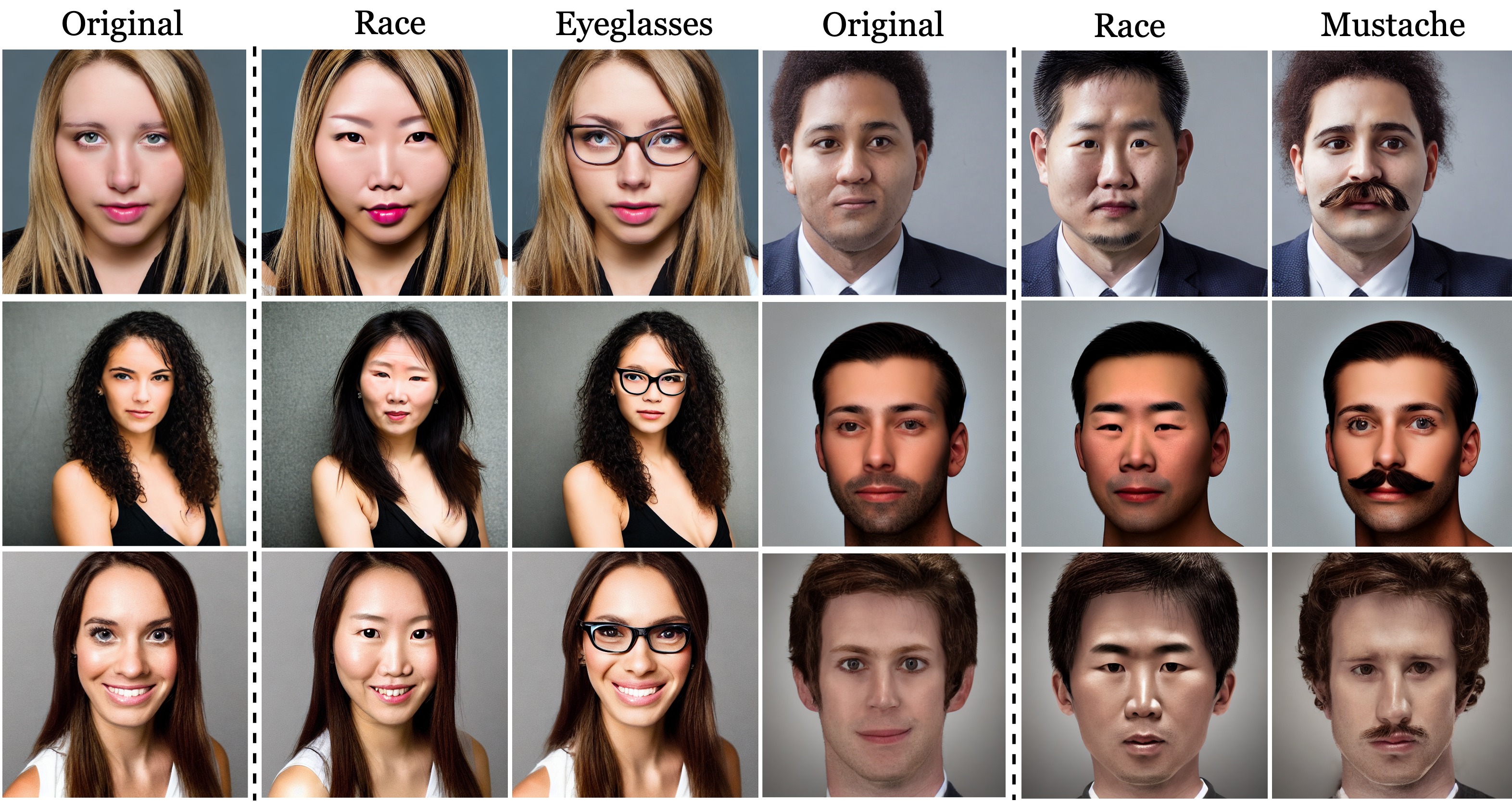}}
    \end{tabular}
  \captionof{figure}{\textbf{Disentagled image editing using \texttt{<EOS>} token.} Original images are displayed on the left side, and the edited versions are on the right. All edits are conducted using the   \texttt{<EOS>} token related to the respective attribute, such as the  \texttt{<EOS>} token of \textit{`A woman with eyeglasses'} or \textit{`A man with mustache'}.}
      \label{fig:teaser}
}] 

\maketitle

\begin{abstract}
Diffusion models have become prominent in creating high-quality images. However, unlike GAN models celebrated for their ability to edit images in a disentangled manner, diffusion-based text-to-image models struggle to achieve the same level of precise attribute manipulation without compromising image coherence. In this paper, CLIP which is often used in popular text-to-image diffusion models such as Stable Diffusion is capable of performing disentangled editing in a zero-shot manner. Through both qualitative and quantitative comparisons with state-of-the-art editing methods, we show that our approach yields competitive results. This insight may open opportunities for applying this method to various tasks, including image and video editing, providing a lightweight and efficient approach for disentangled editing.
 
\end{abstract}
\vspace{-15pt}
\section{Introduction}
 
\begin{figure}
\centering
\begin{subfigure}{0.5\textwidth}
    \includegraphics[scale=0.35]{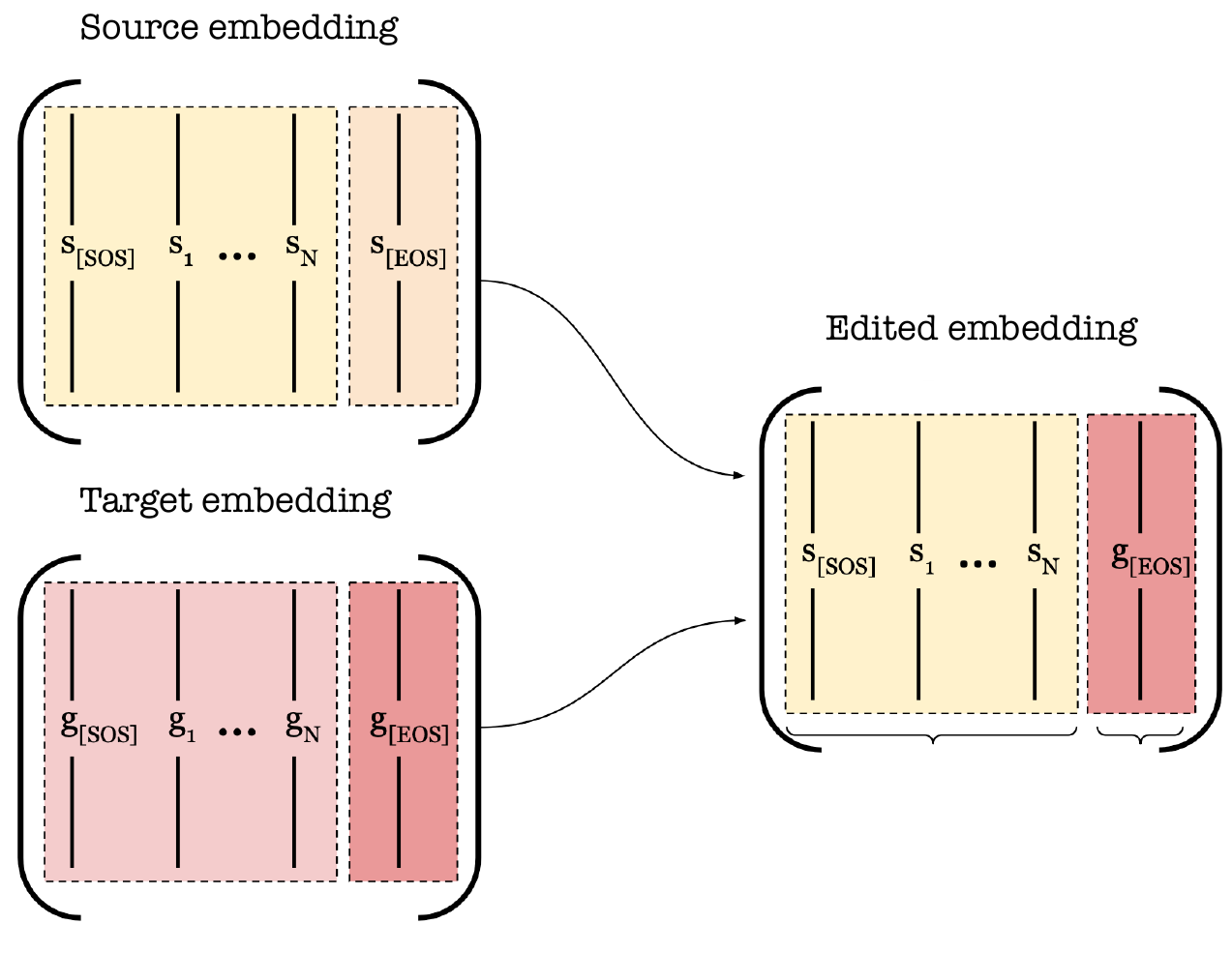}
\end{subfigure}
\caption{Given a source embedding $s$ of a text prompt such `A woman' and a target embedding $g$ such as `A person with an eyeglass', we would like to modify the source embedding $s$ according to $g$ to reflect the corresponding change by replacing  \texttt{<EOS>} token of $g$ with source embedding $s$. }
\label{fig:method}
\end{figure}

Denoising Diffusion Models (\textcolor{black}{DDPMs}) \cite{ho2020denoising} and Latent Diffusion Models (LDMs) \cite{rombach2022high} have garnered significant interest for their capacity to generate high-quality, high-resolution images across various domains. They have marked notable achievements in generative modeling, especially with text-to-image models such as Stable Diffusion \cite{rombach2022high}. 

In generative models, a key aspect of image editing is the disentangled manipulation of semantics, aiming to alter semantically relevant areas of an image without impacting other regions \cite{mathieu2019disentangling, xia2022gan}. Prior studies have shown that \textcolor{black}{latent space disentanglement is achieved easily in GANs as compared to diffusion models}. This has led to considerable research efforts focused on both supervised and unsupervised methods for navigating latent directions in GANs \cite{yuksel2021latentclr, harkonen2020ganspace, shen2020interfacegan}. However, achieving disentangled editing in diffusion models poses a significant challenge. Unlike GANs, which have a structured latent space that naturally lends itself to such disentangled edits, diffusion models operate on a different principle that doesn't inherently support disentangled editing. This is due to the way diffusion models progressively refine images from noise, which complicates the precise control over specific image attributes without affecting others. Despite their impressive capabilities in generating detailed and coherent images, this limitation has been a bottleneck for using diffusion models in tasks that require fine-grained semantic modifications.

Several works have been proposed to achieve disentangled editing in diffusion models, yet they often necessitate expensive procedures such as additional training or fine-tuning \cite{gandikota2023sliders, dalva2023noiseclr, wu2023uncovering}. These approaches, while effective, significantly increase the computational and time costs associated with applying diffusion models for image editing tasks. In this paper, we share an intriguing observation: the CLIP, integral to text-to-image diffusion models, covertly functions as a zero-shot image editing tool. This revelation opens the door to leveraging the   capabilities of the CLIP model embedded within diffusion models for image editing, bypassing the need for costly additional training processes. 

\section{Related work}
\label{sec:related}
 
The utilization of diffusion models for image editing tasks has garnered growing interest within the field of image generation. A prevalent approach is to use text prompts to dictate the desired modifications, but this method often results in entangled edits, where unintended parts of the image are inadvertently altered. Noteworthy exceptions, such as the research conducted by \cite{hertz2022prompt} and \cite{zhang2023adding}, demonstrate more precise editing techniques. For instance, \cite{zhang2023adding}'s ControlNet employs a conditional diffusion model, which permits users to alter specific attributes of an image through conditions. Similarly, studies like \cite{unitune} manage to retain the original content integrity by finely tuning the diffusion model to the input image. Moreover, works by \cite{mokady2023null, han2023improving, wu2023uncovering, huberman2023edit} introduce methods for accurate input image reconstruction, enabling content-preserving edits with classifier-free guidance. While these methods excel in preserving the original appearance during edits, the need for image-specific optimization limits their practicality for instantaneous editing applications.

Recent developments have investigated modifications to the denoising steps of stochastic diffusion models to streamline the editing process. Although these advancements promise more realistic modifications, crafting an optimal editing prompt that maintains the realism and fidelity of the edits to the original image poses a challenge. To address issues of flexibility, some studies, like \cite{brack2023sega} and \cite{liu2022compositional}, suggest decomposing the editing process into several stages. Nonetheless, these methods encounter difficulties when applying multiple edits concurrently, often resulting in compounded effects when various changes are made to the same image. Recent investigations, such as by \cite{dalva2023noiseclr}, have shown success in applying disentangled edits in large-scale models like Stable Diffusion. However, the unsupervised nature of these approaches limits the identification of a comprehensive set of disentangled directions, thus narrowing their adaptability.

\begin{figure}
\centering
\begin{subfigure}{0.5\textwidth}
    \includegraphics[width=0.9\textwidth]{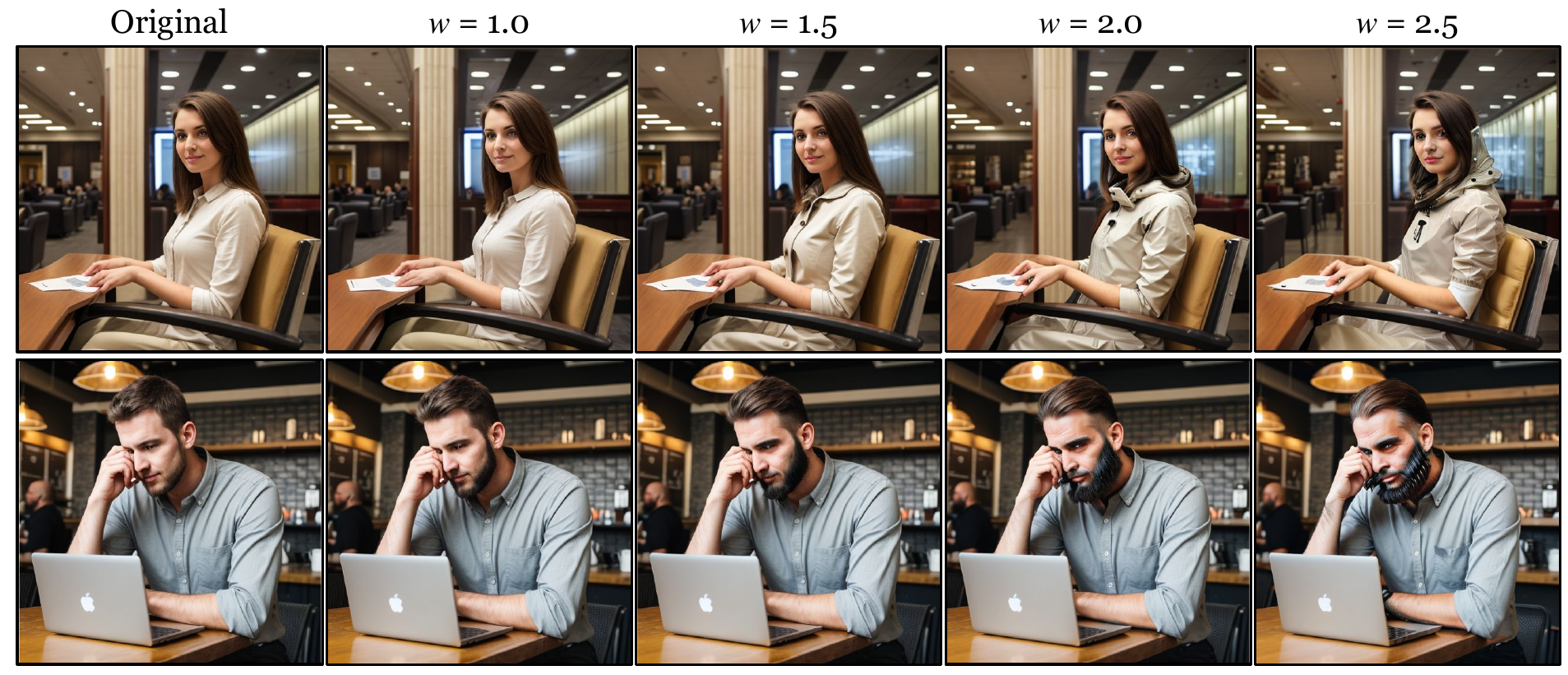}
\end{subfigure}
\caption{\textbf{Target \texttt{<EOS>} Guidance Scale Ablation.} We investigate the trade-off between editing quality vs. preservation depending on the target \texttt{<EOS>} guidance scale hyperparameter.}
\label{fig:hyperparameter_ablation}
\end{figure}

\begin{figure*}
\centering
\begin{subfigure}{1\textwidth}
    \includegraphics[width=1\textwidth]{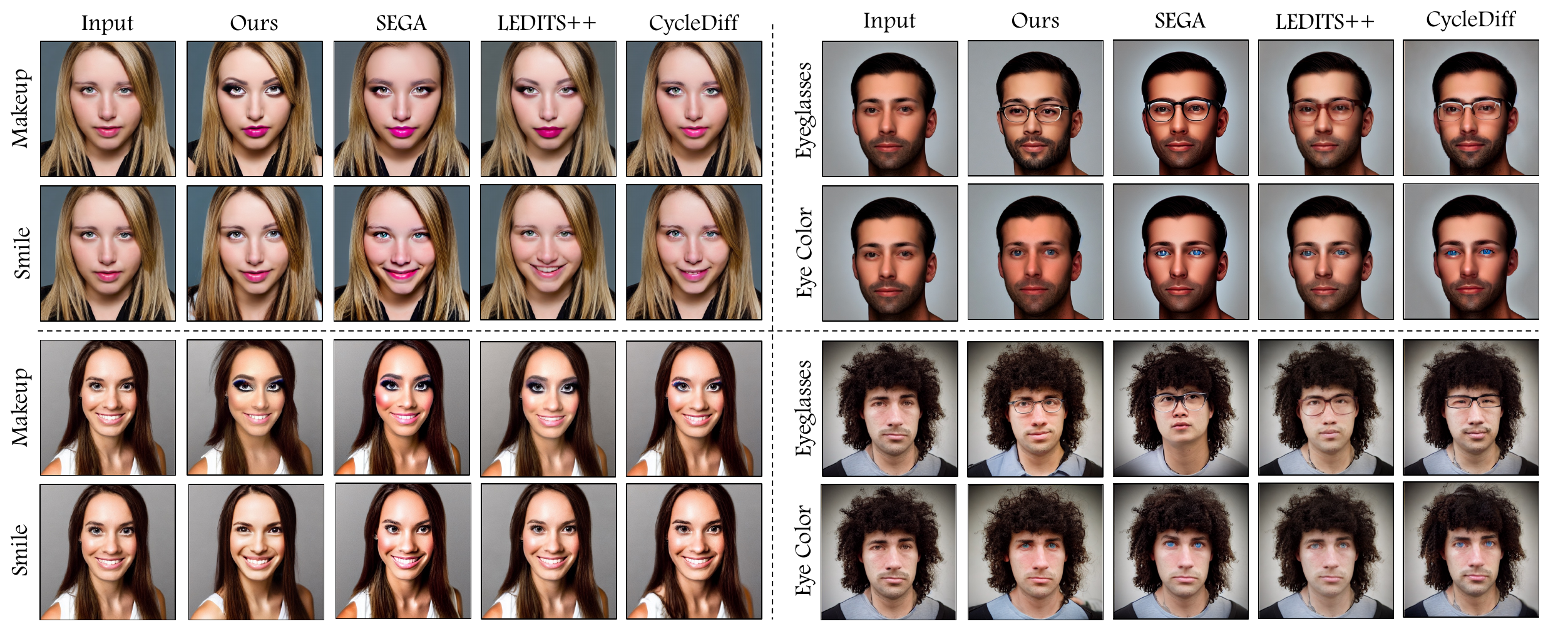}
\end{subfigure}
\caption{\textbf{Qualitative comparison.} We compare the image editing capabilities of \texttt{<EOS>} token with state-of-the-art methods  SEGA \cite{brack2023sega}, Ledits++ \cite{brack2023ledits++} and Cycle Diffusion \cite{cyclediffusion}. }
\label{fig:appendix_face_man}
\hfill
\label{fig:comparison}
\end{figure*}
% \lipsum[2-8]

\section{Methodology}

Given a source embedding $s$ of a text prompt such `A woman' and a target embedding $g$ such as `A person with an eyeglass', we would like to modify the source embedding $s$ according to $g$ to reflect the corresponding change.   Our method is inspired by a novel observation: \texttt{<EOS>} token in CLIP model is capable of performing disentangled edits in a zero-shot manner.   By leveraging the fact that CLIP has a decoder-only text encoder and performs a causal language encoding, we can utilize the \texttt{<EOS>} representation of the target embedding $g$ to change the context of the source embedding $s$.  Thus, we define $\sigma(s, g)$ for arbitrary text conditions $s$ and $g$  as follows:

\begin{equation}
    \sigma(s, g) = [s_{\text{\texttt{<SOS>}:N}} \mid w \times g_{\texttt{<EOS>}}],
\end{equation}
in which $s_{\text{[SOS]:N}} \in \mathbb{R}^{d \times (N+1)}$ where \texttt{<SOS>} represents start of the sentence token, $g_{\text{[EOS]}} \in \mathbb{R}^{d \times 1}$, and $w$ is the controllable target \texttt{<EOS>} guidance hyperparameter.  Then given a source embedding $s$, such as \textit{a man}, and a guidance prompt $g$, such as \textit{a person with mustache}, we then employ $\gamma = \sigma(s, g)$ and generate the modified image  using edited embedding  $\gamma$. See Fig. \ref{fig:method} for an illustration of the editing operation.  This reformulation of the source embedding eliminates the necessity for unsupervised training of new tokens or finetuning.

\begin{figure*}
\centering
\begin{subfigure}{1\textwidth}
    \includegraphics[width=1\textwidth]{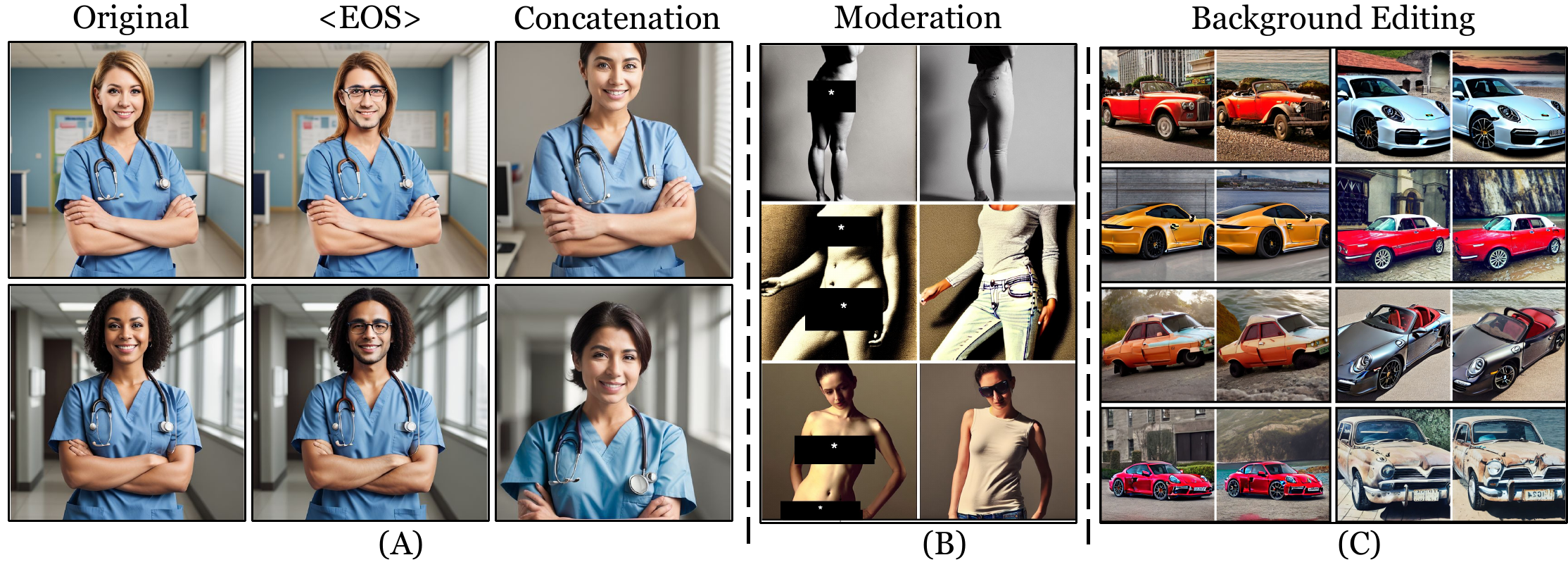}
\end{subfigure}
\caption{\textbf{Qualitative comparison.} (A) We compared our method with a prompt concatenation baseline. We used "a nurse" as the source prompt, and "man, glasses" as the target prompt for generating \texttt{<EOS>} guided image editing. The prompt for the baseline is "a nurse, man, glasses". All images were generated with identical initial noise. \textcolor{black}{High-quality generation was facilitated by Realistic Vision V6.} (B) \texttt{<EOS>}-based editing can be used for content moderation. (C) \texttt{<EOS>}-based editing is an effective technique for background editing as well.}
\hfill
\label{fig:combined_exps}
\end{figure*}
\vspace{-5pt}

\section{Experiments}

We demonstrate the efficiency of image editing within text-to-image diffusion models by employing the \texttt{<EOS>} token across diverse scenarios, such as editing images of faces, cars, and moderating nude content. Additionally, we benchmark these edits against state-of-the-art editing methods and conduct a user study to quantitatively assess our technique. \textcolor{black}{We used SD 1.4 for our experiments. To generate (StableDiffusion - \texttt{<EOS>}) paired images at inference time, we use the same noise under the same seed. Next, we obtain the text embeddings of source prompt (such as "a nurse") and target prompt (such as "man with eyeglasses"). Finally, we swap the \texttt{<EOS>} embedding of the source prompt with the \texttt{<EOS>} embedding of the target prompt. In all experiments, we used 50 steps of denoising.}

\vspace{-15pt}
 \paragraph{Qualitative Comparison} We compare 
 \texttt{<EOS>}-based editing  with state-of-the-art image editing methods SEGA \cite{brack2023sega}, Ledits++ \cite{brack2023ledits++} and Cycle Diffusion \cite{cyclediffusion}. For every attribute of interest, like \textit{eyeglasses}, we process it using the CLIP text encoder to acquire the \texttt{<EOS>} embedding of that specific attribute. Then,   $\sigma(s, g) = [s_{\text{\texttt{<SOS>}:N}} \mid g_{\text{\texttt{<EOS>}}}]$ is employed at inference time, where the source prompt is \textcolor{purple}{a headshot of a woman} or \textcolor{purple}{a headshot of a man} (see  Fig. \ref{fig:teaser} and  Fig. \ref{fig:comparison}). When we combine the \texttt{<EOS>} embedding of a specific attribute with the source prompt, like \textit{a headshot of a woman}, we are essentially editing the underlying concept in the source prompt. Subsequently, all images generated with the same source prompt begin to exhibit visual elements that align with the integrated attribute. As can be seen from Fig. \ref{fig:comparison}, our method achieves comparable results with state-of-the-art methods such as SEGA \cite{brack2023sega}, Ledits++ \cite{brack2023ledits++} and Cycle Diffusion \cite{cyclediffusion}.

\vspace{-5pt}
\paragraph{Complex edits}  We also test more complex edits such as altering the background of an image. For this purpose, we take random car pictures obtained by SD-1.4 and fuse them  with  \texttt{<EOS>}  embedding of text prompt \textit{sea}. In Fig. \ref{fig:combined_exps}\textcolor{red}{.C} we demonstrate that while the composition of car structures remains consistent, the successful integration of sea information in the \texttt{<EOS>}  embedding leads to a notable transformation in the scene background, effectively replacing it with a seascape. We also demonstrate the trade-off between content preservation and edit quality in Fig.\ref{fig:hyperparameter_ablation}.
% \vspace{-5pt}
\begin{table}[]
    \centering
    \begin{tabular}{@{}lcc@{}}
        \toprule
         \textbf{Method} &  Edit Quality & Disentanglement \\ \hline
         SEGA \cite{brack2023sega} & 2.76 & 2.64 \\
         Cycle Diff. \cite{cyclediffusion} & 2.99 & 2.99 \\
         LEDITS++ \cite{brack2023ledits++} & \textbf{3.58} & \textbf{3.27} \\ \hline
         Ours & \underline{3.20} & \underline{3.12} \\ \hline
    \end{tabular}
    \caption{\textbf{User Study Results.} The average user responses are provided in the table. We conduct our study within a range of 1-to-5.}
    \label{tab:user_study}
\end{table}

\paragraph{Moderating NSFW Content} We illustrate the effectiveness  of \texttt{<EOS>} guidance during the inference stage through its application in a NSFW (Not Safe for Work) moderation scenario (see Fig.\ref{fig:combined_exps}\textcolor{red}{.B}). By incorporating \texttt{<EOS>} into the moderation process by $\sigma$(\textcolor{purple}{\texttt{<NSFW>}}, \textcolor{purple}{dressed $\texttt{<gender>}$}), we aim to demonstrate its capability to enhance content filtering and ensure a more secure and appropriate online environment. This application not only highlights the technical prowess of \texttt{<EOS>} but also emphasizes its potential impact in real-world contexts where sensitive content moderation is crucial. As can be seen from the results, our method can successfully moderated unsafe content while keeping the original structure of the images preserved.  
\vspace{-10pt}
\paragraph{\textcolor{black}{Mean Opinion Score (MOS).}} We conduct a user study with 25 participants on the Prolific platform\footnote{\url{https://www.prolific.com}} to evaluate the effectiveness of our method in terms of edit quality and disentanglement capabilities. The participants were shown a series of input-edit pairs and asked to evaluate them on whether the intended edit has been applied successfully and whether the identity of the input is preserved. For each of the questions, the participants are asked to assign a rating within the scale of 1-to-5 where 5 corresponds to the highest score. Referring to the results demonstrated in Table \ref{tab:user_study}, our method outperforms both \cite{brack2023sega} and \cite{brack2023ledits++} and performs comparably with \cite{brack2023ledits++}.

\section{Limitations}
\textcolor{black}{Biases within the CLIP can affect editing  results, an issue suffered by SOTA editing methods as well. For example, when  altering a person’s eye color to blue, even SOTA methods may inadvertently remove the beard (see the last row of Fig.\ref{fig:comparison}). This behavior inherently affects the disentanglement capability of the proposed method.}

\section{Conclusion}
This paper has illuminated the previously unexplored territory of CLIP's capabilities as a zero-shot image editing method, leveraging its EOS token to bridge the gap between textual prompts and visual outputs. 
\newpage
 
\bibliographystyle{ieeenat_fullname}
\bibliography{main}

% WARNING: do not forget to delete the supplementary pages from your submission
\newpage
\clearpage
\makeatletter
\renewcommand \thesection{S.\@arabic\c@section}
\renewcommand\thetable{S.\@arabic\c@table}
\renewcommand \thefigure{S.\@arabic\c@figure}
\makeatother
\setcounter{section}{0}
\setcounter{page}{1}
\maketitlesupplementary

\section{Additional results}
\begin{figure}[t]
\centering
\begin{subfigure}{0.5\textwidth}
    \includegraphics[width=1\textwidth]{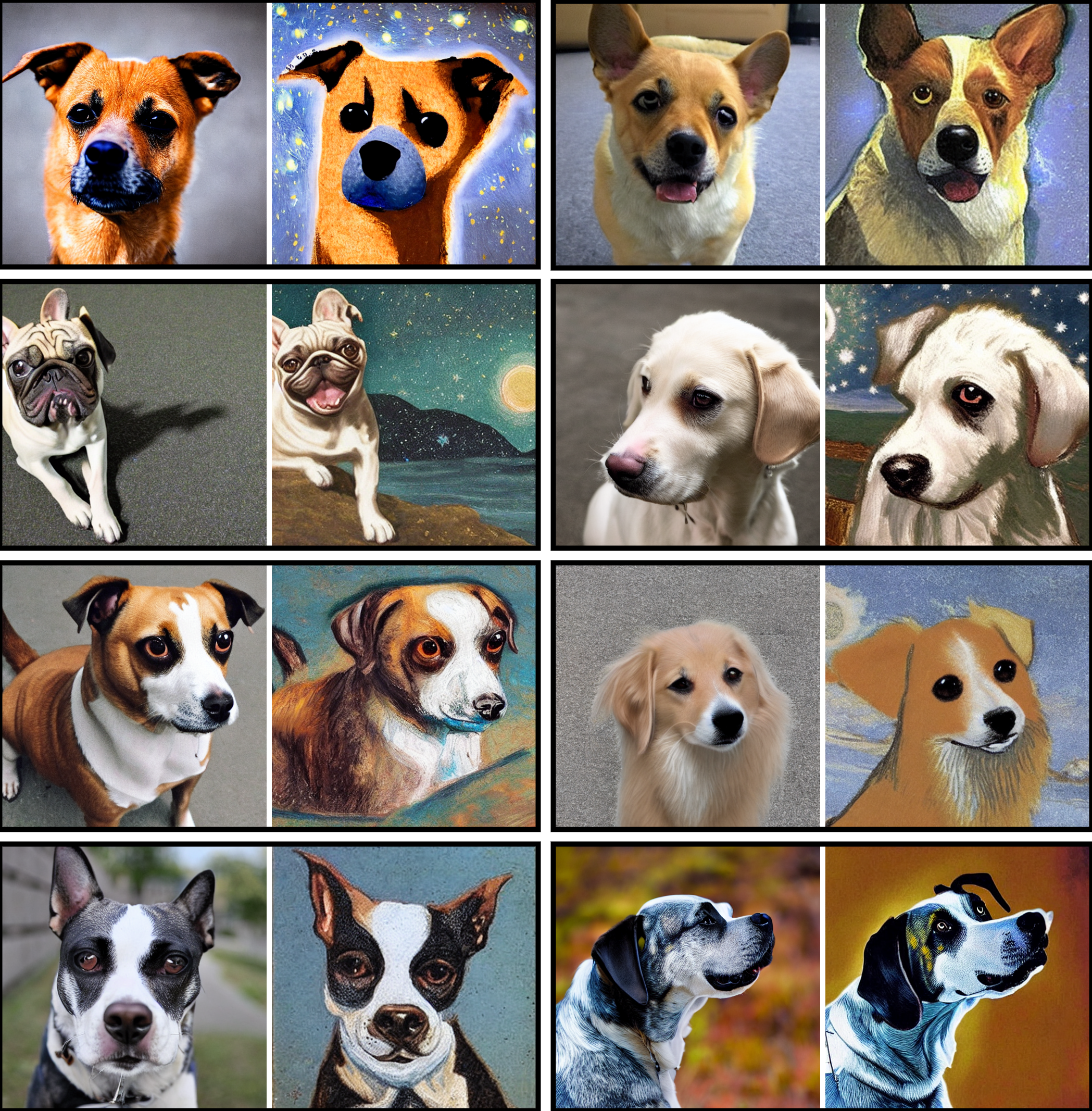}
\end{subfigure}
\caption{We show the editing capabilities of [EOS] guidance beyond marginal attributes. We change the the entire theme to a painting theme. All pictures are generated with the text prompt \textcolor{purple}{a dog}. In order to obtain the edited versions, $\sigma$(\textcolor{purple}{a dog}, \textcolor{purple}{painting}) is applied.}
\label{fig:appendix_marginal_dog}
\hfill
\label{fig:figures}
\end{figure}

Please see Figure \ref{fig:appendix_marginal_dog}

\section{Details about User Study}
We ask the following questions to the users: 

\begin{itemize}
    \item For edit quality: \textit{The original image is shown on the left, and the modified image is shown on the right. How likely do you think the modified image is depicting the same person while featuring "Makeup" attribute? Rate from 1 (Not at all) to 5 (Very well)}
\item  For disentanglement: \textit{The original image is shown on the left, and the modified image is shown on the right. How likely do you think the modified image reflects "Makeup" feature? Rate from 1 (Not at all) to 5 (Very well)}
\end{itemize}

\end{document}